\pdfoutput=1
\documentclass[10pt,twocolumn,letterpaper]{article}

\usepackage{cvpr}
\usepackage{times}
\usepackage{epsfig}
\usepackage{graphicx}
\usepackage{amsmath}
\usepackage{amssymb}
\usepackage{multirow}
\usepackage{diagbox}
\usepackage{booktabs}

\usepackage[pagebackref=true,breaklinks=true,colorlinks,bookmarks=false]{hyperref}

\cvprfinalcopy 

\def\cvprPaperID{4324} 

\ifcvprfinal\fi
\begin{document}

\title{VESR-Net: The Winning Solution to Youku Video Enhancement and Super-Resolution Challenge}

\author{Jiale Chen\\
University of Science and Technology of China\\
{\tt\small chenjlcv@mail.ustc.edu.cn}
\and
Xu Tan\\
Microsoft Research Asia\\
{\tt\small xuta@microsoft.com}
\and
Chaowei Shan\\
University of Science and Technology of China\\
{\tt\small cwshan@mail.ustc.edu.cn}
\and
Sen Liu\\
University of Science and Technology of China\\
{\tt\small elsen@iat.ustc.edu.cn}
\and
Zhibo Chen\\
University of Science and Technology of China\\
{\tt\small chenzhibo@ustc.edu.cn}
}

\maketitle

\begin{abstract}

This paper introduces VESR-Net, a method for video enhancement and super-resolution (VESR). We design a separate non-local module to explore the relations among video frames and fuse video frames efficiently, and a channel attention residual block to capture the relations among feature maps for video frame reconstruction in VESR-Net. We conduct experiments to analyze the effectiveness of these designs in VESR-Net, which demonstrates the advantages of VESR-Net over previous state-of-the-art VESR methods. It is worth to mention that among more than thousands of participants for Youku video enhancement and super-resolution (Youku-VESR) challenge, our proposed VESR-Net beat other competitive methods and ranked the first place.


\end{abstract}

\section{Introduction}
Video enhancement and super-resolution (VESR)~\cite{kim20183dsrnet,hyun2018spatio,xue2019video,liu2013bayesian,wang2019edvr} aims to recover high-resolution details from noisy and low-resolution video frames, and has draw a lot of attention both in the research and industrial community. Recent research on VESR largely focuses on some public datasets~\cite{zhang2018image,nah2019ntire}, which usually use bicubic down-sampling~\cite{gavade2014super} to obtain the low-resolution videos and cannot cover the degradations in industrial scenarios~\cite{stuhlmuller2000analysis,wiegand2003overview}. In order to push the VESR research towards restoring high-quality videos from low-quality videos that suffer from real-world degradations, Youku hosted the video enhancement and super-resolution (Youku-VESR) challenge\footnote{The challenge webside is here: \url{https://tianchi.aliyun.com/competition/entrance/231711/rankingList/1}} to explore the VESR solutions on the datasets that cover the realistic degradations in Youku online video watching application~\footnote{\url{https://www.youku.com/}}. 

We introduce our participated solution, VESR-Net, which won the first place in Youku-VESR challenge. Specifically, we design a \textit{separate non-local (Separate NL)} module to explore the relations among video frames and fuse video frames efficiently, and a \textit{channel attention residual block (CARB)}~\cite{zhang2018image} to capture the relation among feature maps for video frame reconstruction in VESR-Net. Compared with previous VESR methods~\cite{zhang2018image,wang2019edvr} that suffer from high computation complexity, VESR-Net is more efficient while more effective.

In the following sections, we first briefly introduce Youku-VESR challenge, and describe the related works on VESR. We then introduce our proposed VESR-Net in details and conduct experimental studies to verify the effectiveness of designs in VESR-Net. We further show the results of VESR-Net in Youku-VESR challenge.
 

\begin{figure*}
\small
  \centering
  \includegraphics[width=1.0\textwidth]{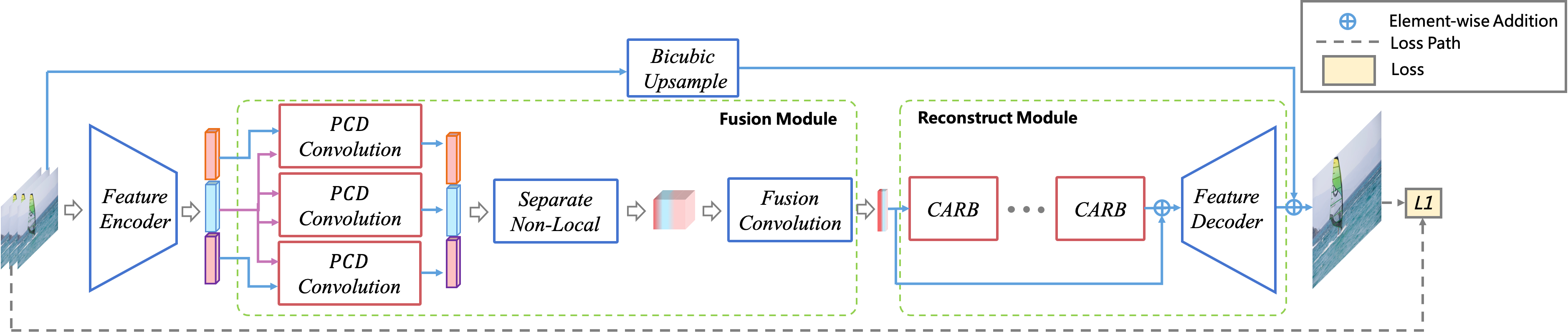}
  \caption{\small The overview of VESR-Net. We employ both PCD convolution in \cite{wang2019edvr} and separate non-local architecture to aggregate the information among different frames. For reconstruction, we utilize stacked channel-attention residual block (CARB)~\cite{zhang2018image} followed with a feature decoder. The gray arrow denotes the L1 loss applied on the central target frame.}
  \label{framework}
\end{figure*}


\section{Related Work}

Video enhancement and super-resolution has attracted more and more research attention in recent years. How to explore the relations between adjacent video frames is essential to achieve good performance. Recently, many progresses of research were made~\cite{caballero2017real,haris2019recurrent,hyun2018spatio,kim20183dsrnet,sajjadi2018frame,tao2017detail,wang2019edvr}. RCAN~\cite{zhang2018image} is the state-of-the-art algorithm for image super-resolution. ~\cite{li2019fast,kim20183dsrnet} proposed 3-dimensional (3D) convolution neural networks to exploit the spatial and temporal information of video frames simultaneously. Li \etal. \cite{li2019fast} proposed 3-dimensional (3D) convolution neural networks to exploit the spatial and temporal information of video frames simultaneously. Kim~\etal \cite{kim20183dsrnet} proposed a spatio-temporal network to mitigate occlusions in optical flow estimation. Xue \etal. \cite{xue2019video} proposed a framework to exploit the temporal relation with the optical flow. Recent methods \cite{haris2019recurrent,sajjadi2018frame} pay attention to recurrent framework which is efficient to learn the sequential features. Jo \etal. \cite{jo2018deep} designed a dynamic upsampling filters for video super-resolution which are computed depending on the local spatio-temporal neighborhood of each pixel to avoid explicit motion compensation. 

However, most algorithms suffer from high computation complexity. For example, RCAN~\cite{zhang2018image} ignores the cross-frame information and needs to stack more than 400 layers to achieve good reconstruction accuracy. EDVR~\cite{wang2019edvr} fuses cross-frame information with deformable network and attention mechanism, which still needs to stack 40 residual blocks for final reconstruction. Different from these methods, our method aims to explore the relations among different frames with attention mechanism to improve the information fusion among frames and leverage the residual attention module to improve the frame reconstruction.

\section{Youku-VESR Challenge}
The objective of the Youku-VESR challenge is to push the state-of-the-art VESR methods towards restoring low-quality videos with realistic degradations. Youku-VESR challenge collects 1000 1080p video clips, with both high-resolution and low-resolution video pairs. This dataset contains diverse contents that cover a variety of categories, and the low-resolution videos suffer from different noises incurred in the online video watching application. 

\paragraph{Challenge phases} (i) \textit{The first phase: } All participants got 200 pairs of low-resolution (LR) and high-resolution (HR) videos for training and 50 LR videos for evaluation. An online validation server with a leaderboard provided accuracy score for the uploaded HR video results corresponding to the LR videos in the evaluation set. (2) \textit{The second phase: } Youku releases 650 pairs of LR and HR videos for training and 100 LR videos for validation. The LR videos in the second phase suffer from severer degradations than those in the first phase.

\paragraph{Evaluation phases} The quantitative metrics are Peak Signal-to-Noise Ratio (PSNR) measured in deciBels [dB] and video multi-method assessment fusion (VMAF)~\cite{li2016toward}. These metrics are computed between the generated HR results and the ground-truth HR video frames. Each solution will be evaluated on 50 videos provided by Youku. They tested all video frames in top 5 videos and tested one frame in every 5 frames for remaining videos. 

\begin{figure*}
  \centering
  \includegraphics[width=1.0\textwidth]{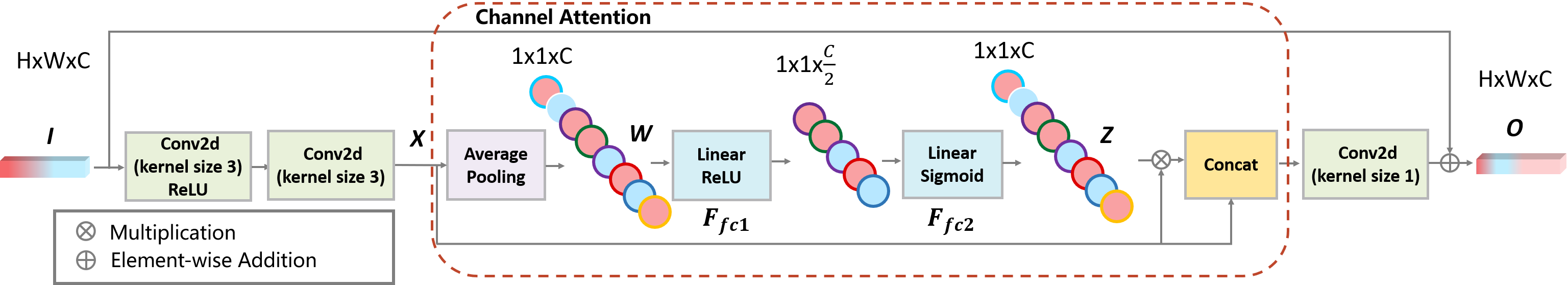}
  \caption{\small The architecture of Channel Attention Residual Block.}
  \label{CA}
\end{figure*}

\section{Methods}
In this section, we introduce our method used in Youku-VESR challenge, which is called as VESR-Net.

\subsection{VESR-Net}
In VESR, the information across frames and across pixels in each frame could be useful for the reconstruction of high-quality frame. However, the disturbing frames need to be ignored by capturing the relation among different frames, due to that scene switching is common among different video frames. We also need to selectively focus on other pixels in the same frame for reconstruction. Therefore, how to fuse the information across video frames and across pixels in each frame is the key to achieve good performance, which has not been exploited well by previous works. Therefore, we propose VESR-Net to enhance the inter-frame and intra-frame fusion for high-quality video frame reconstruction. As shown in Figure~\ref{framework}, VESR-Net consists of two components: a fusion module with inter-frame fusion and a reconstruct module with intra-frame fusion. In the fusion module, we aim to fuse neighboring frames for middle frame reconstruction. In this module, we explore the useful information from neighboring frames and at the same time ignore the temporal redundancy information. Therefore, we propose a \textit{Separate Non-Local (Separate NL)} architecture to model the relation among video features. In frame reconstruction module, we introduce channel-wise attention mechanism in residual block (denoted by CARB) for efficient reconstruction. We describe the details of \textit{Separate Non-Local (Separate NL)} in fusion module and \textit{Channel-Attention Residual Block (CARB)} in reconstruction module in the following subsections. The detailed configurations of VESR-Net is shown in Table~\ref{network}.

\begin{figure}[h!]
\small
  \centering
  \includegraphics[width=0.5\textwidth]{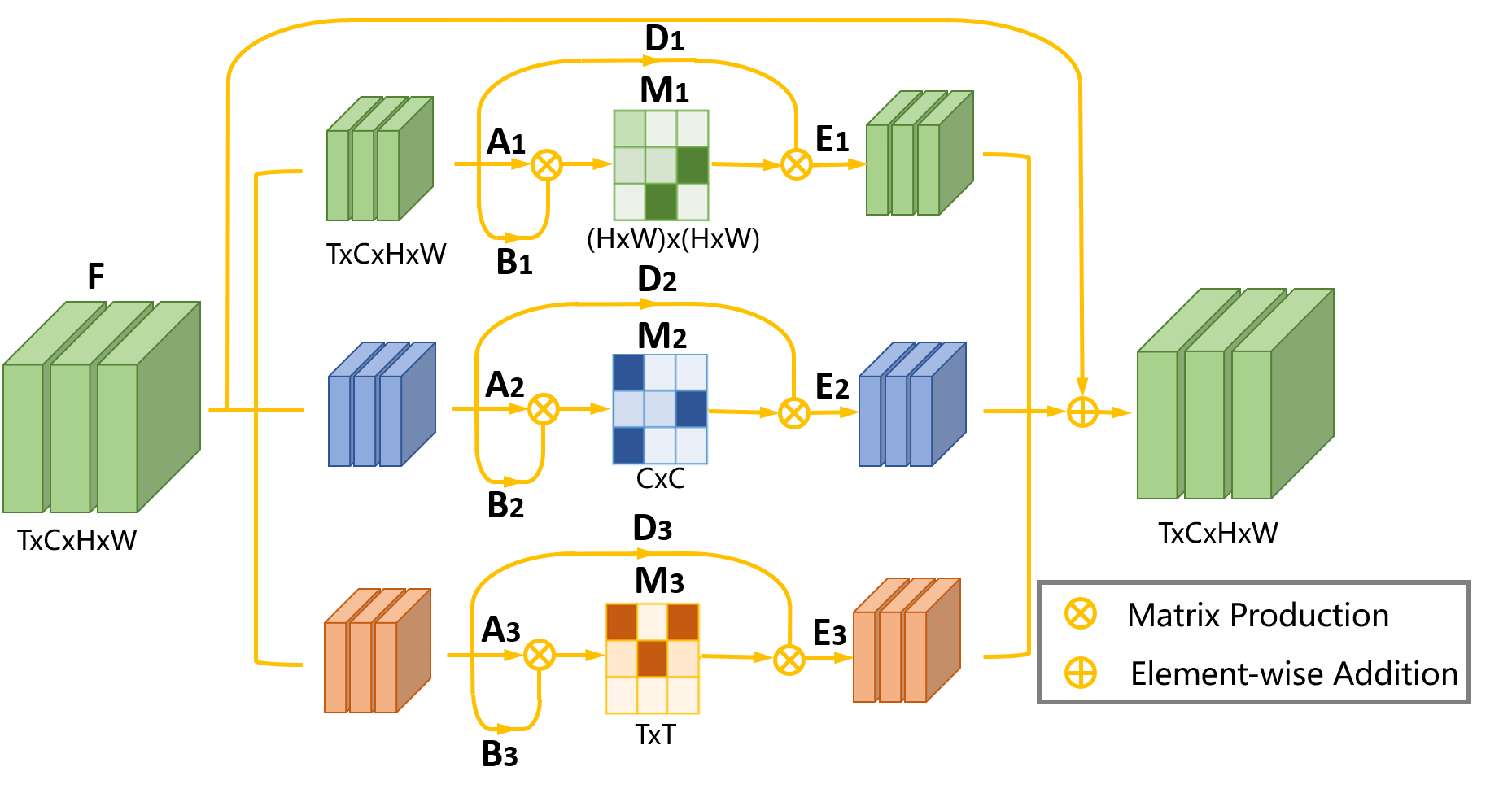}
  \caption{\small The architecture of Separate Non-local.}
  \label{separate_nl}
\end{figure}

\newcommand{\tabincell}[2]{\begin{tabular}{@{}#1@{}}#2\end{tabular}}

\begin{table*}[t]\footnotesize
\begin{center}
\begin{tabular}{c|c|c|c|c|c|c|c}
\toprule
\multicolumn{2}{c|}{{\bf VESR-Net}} & Input & Output & In channels & Out channels & Kernel size & Stride \\
\midrule
\multirow{2}{*}{\tabincell{c}{Feature\\Encoder}} & Conv1 & ($X_{t-3}$,$\cdots$,$X_{t}$,$\cdots$,$X_{t+3}$) & ($f^{1}_{t-3}$,$\cdots$,$f^{1}_{t}$,$\cdots$,$f^{1}_{t+3}$) & 3 & 128 & 3$\times$3 & 1  \\
& CARB2-6 & ($f^{1}_{t-3}$,$\cdots$,$f^{1}_{t}$,$\cdots$,$f^{1}_{t+3}$) & ($f^{6}_{t-3}$,$\cdots$,$f^{6}_{t}$,$\cdots$,$f^{6}_{t+3}$) & 128 & 128 & - & -  \\
\midrule
\multirow{3}{*}{\tabincell{c}{Fusion\\Module}}& PCD Conv7~\cite{wang2019edvr} & ($f^{6}_{t-3}$,$\cdots$,$f^{6}_{t}$,$\cdots$,$f^{6}_{t+3}$) & ($f^{7}_{t-3}$,$\cdots$,$f^{7}_{t}$,$\cdots$,$f^{7}_{t+3}$) & 128 & 128 & - & -  \\
& Seperate NL8 & ($f^{7}_{t-3}$,$\cdots$,$f^{7}_{t}$,$\cdots$,$f^{7}_{t+3}$) & ($f^{8}_{t-3}$,$\cdots$,$f^{8}_{t}$,$\cdots$,$f^{8}_{t+3}$) & 128 & 128 & - & -  \\
& Conv9 & ($f^{8}_{t-3}$,$\cdots$,$f^{8}_{t}$,$\cdots$,$f^{8}_{t+3}$) & $F^{9}_{t}$ & 896 & 128 & 3$\times$3 & 1  \\
\midrule
\multirow{7}{*}{\tabincell{c}{Reconstruct\\Module}} & CARB10-30 & $F^{9}_{t}$ & $F^{30}_{rec}$ & 128 & 128 & - & -  \\
& Conv31 & $F^{30}_{rec}$ & $F^{31}_{rec}$ & 128 & 512 & 3$\times$3 & 1  \\
& PixelShuffle32~\cite{shi2016real} & $F^{31}_{rec}$ & $F^{32}_{rec}$ & 512 & 128 & - & -  \\
& Conv33 & $F^{32}_{rec}$ & $F^{33}_{rec}$ & 128 & 256 & 3$\times$3 & 1  \\
& PixelShuffle34~\cite{shi2016real} & $F^{33}_{rec}$ & $F^{34}_{rec}$ & 256 & 64 & - & -  \\
& Conv35 & $F^{34}_{rec}$ & $F^{35}_{rec}$ & 64 & 64 & 3$\times$3 & 1  \\
& Conv36 & $F^{35}_{rec}$ & $H_{t}$ & 64 & 3 & 3$\times$3 & 1  \\
\bottomrule
\end{tabular}
\end{center}
\caption{The configurations of VESR-Net. Conv denotes the convolution layer, CARB denotes the channel-attention residual block. ($X_{t-3}$,$\cdots$,$X_{t}$,$\cdots$,$X_{t+3}$) denote the input video frames}
\label{network}
\end{table*}

\begin{figure*}
\small
  \centering
  \includegraphics[width=0.9\textwidth]{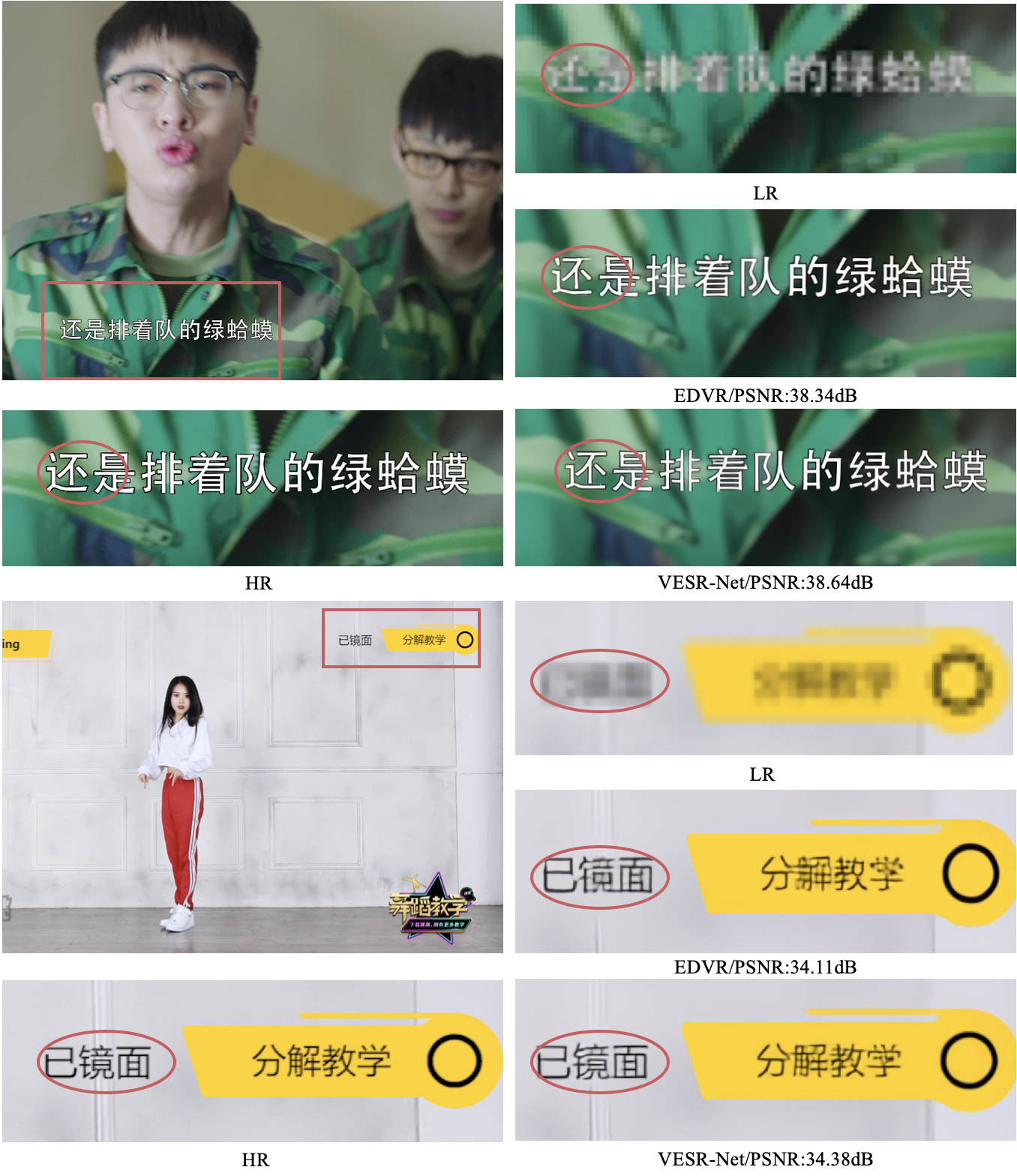}
  \caption{\small The qualitative results of EDVR and VESR-Net on Youku-VESR challenge dataset, where ``LR'' and ``HR'' represent the low-resolution and high-resolution frames respectively.}
  \label{qualitative_res}
\end{figure*}

\subsection{Separate Non Local}
Self-attention is widely used in natural language processing (NLP) tasks \cite{vaswani2017attention,devlin2018bert}. Wang \etal. \cite{wang2018non} introduce this mechanism into computer vision architecture called \textit{Non-Local Neural Network}. Given a feature ${\bf F} \in \mathbb{R}^{T \times H \times W \times C}$, we first feed it into a convolution layer to generate two new feature maps ${\bf A}$ and ${\bf B}$. Then we reshape them to $\mathbb{R}^{C \times N}$ where $N = T \times H \times W$ is the number of pixels in video clips. After that, we perform a matrix multiplication between the transpose of ${\bf A}$ and ${\bf B}$, and apply a softmax layer to calculate the attention map ${\bf M} \in \mathbb{R}^{N \times N}$:

\begin{equation}\label{non_local_softmax}
{\bf M}_{ij}=\frac{exp({\bf A}^T \cdot {\bf B})}{\sum_{i=1}^{N} exp({\bf A}^T \cdot {\bf B})}
\end{equation}

\noindent
where $M_{ij}$ measures the impact of the $i_{th}$ position on the $j_{th}$ position. We call it \textit{relation matrix}, which measures the similarity between every two pixels in video features. More similar feature representations between the two position indicates more correlation between them. Therefore, the non-local operation can apply the attention to all pixels in feature map, which is suitable to fuse adjacent frames and ignore the redundant frames in video clips. Meanwhile, we generate a new feature map ${\bf D} \in \mathbb{R}^{C \times H \times W}$ and reshape it to $\mathbb{R}^{C \times N}$. Then we perform a matrix multiplication between {\bf M} and the transpose of {\bf D} and reshape the result to $\mathbb{R}^{C \times H \times W}$.

However, the non-local operation consumes large amount of parameters because of the high dimension of relation matrix ${\bf M} \in \mathbb{R}^{N \times N}$, especially for video features. Therefore, we design a new architecture called \textit{Separate Non-Local} to fuse the information across video frames and across pixels in each frame, which can reach better performance with shallower network. As shown in Figure~\ref{separate_nl}, we design three types of attention modules to explore global contextual information in different dimension. First, we generate two new feature maps ${\bf A_1, A_2, A_3}$ and ${\bf B_1, B_2, B_3}$ in three branches respectively. Then we reshape them to $\mathbb{R}^{C \times T \times N_1}, \mathbb{R}^{T \times H \times W \times N_2}, \mathbb{R}^{C \times H \times W \times N_3}$ where $N_1 = H \times W, N_2 = C, N_3 = T$. And then we perform the matrix production to get three relation matrices ${ {\bf M_1} \in \mathbb{R}^{HW \times HW}, {\bf M_2} \in \mathbb{R}^{C \times C}, {\bf M_3} \in \mathbb{R}^{T \times T} }$. ${\bf M_1}$, ${\bf M_2}$ and ${\bf M_3}$ denote the similarity between different spatial contexts, different channels, and different time steps, respectively. 

Meanwhile, we feed video feature ${\bf F}$ into three convolution layers to generate the new feature map ${\bf D_1, D_2, D_3}$. Then we reshape them to $\mathbb{R}^{C \times T \times N_1}, \mathbb{R}^{T \times H \times W \times N_2}, \mathbb{R}^{C \times H \times W \times N_3}$ respectively. Next, we perform a matrix multiplication between the transpose of ${\bf D_1, D_2, D_3}$ and ${\bf M_1, M_2, M_3}$ to get the results ${\bf E_1, E_2, E_3}$. Finally, we perform an element-wise sum operation between ${\bf E_1, E_2, E_3}$ and ${\bf F}$. At last, we sum up the features from three branch to get the fusion features.

\subsection{Channel Attention Residual Block}
The channel-wise attention mechanism in residual block (CARB) in the reconstruction module is essential for efficient reconstruction and good performance of VESR. As shown in Figure~\ref{CA}, in CARB, we first perform global average pooling to obtain channel descriptors {\bf W}. Then, we obtain the channel weights {\bf Z} with two linear layers and perform multiplication between channel weights {\bf Z} and video features {\bf X}. At last we concatenate the video features with attention weights and obtain the final output with a 1$\times$1 convolution layer.




\section{Experiments and Results}

In this section, we introduce the implement details of VESR-Net in Youku-VESR challenge and compare VESR-Net with state-of-the-art algorithms.

\subsection{Training Setup}
\label{sec_data_setup}
\paragraph{Datasets.} We train and evaluate our method on the dataset used in the second phase of challenge, which contains 1000 video clips. We split 50 videos for evaluation and use the remaining videos for training.

\paragraph{Implementation Details} In our proposed VESR-Net, we adopts five channel attention residual blocks (CARB) as feature encoder to perform feature extraction. We design two VESR-Net with different model sizes: \textit{VESR-Net$_{small}$} with 20 CARBs and \textit{VESR-Net} with 40 CARBs in the reconstruction module, respectively. The channel size in each residual block is set to 128. In all our experiments, we focus on 4$\times$ super-resolution factor. We use video frame patches of size 64$\times$64 as inputs and 256$\times$256 as outputs in training phase but use the original frame size while testing. The model takes 7 consecutive frames as inputs to reconstruct the middle frame. We implement VESR-Net on PyTorch and train the model with 4 NVIDIA Titan 1080Ti GPUs, each with batch size of 32. We use Adam optimizer~\cite{kingma2014adam} with $\beta_1$=0.9 and $\beta_2$=0.999. The initial learning rate is set to $1\times10^{-4}$ and the weight decay parameter is 0.8 for each 20 epoches. Since EDVR~\cite{wang2019edvr} is the state-of-the-art algorithms in VESR, we compare our method with EDVR on the dataset used in Youku-VESR challenge. We reproduce EDVR following the settings in the original paper and released code~\cite{wang2019edvr}. For fair comparison, \textit{EDVR$_{small}$} adopts 20 residual block, which has the comparable amount of parameters with \textit{VESR-Net$_{small}$}.

\begin{table}[t]\scriptsize
\begin{center}
\begin{tabular}{c||c|c|c|c}
\hline
 & EDVR$_{small}$ & EDVR & VESR-Net$_{small}$ & VESR-Net \\
\hline
PSNR  & 35.47 & 35.75 & {\bf 35.88} & {\bf 35.97} \\
FLOPs & 152.93G & 177.11G & 162.14G & 186.26G \\
Params & 14.80M & 20.7M & 15.96M & 21.65M \\
\hline
\end{tabular}
\end{center}
\caption{The  quantitative evaluation results on dataset used in the second phase of Youku-VESR Challenge. FLOPs \cite{molchanov2016pruning} are calculated on low-resolution video frames with the size of 64 $\times$ 64.}
\label{experiment}
\end{table}

\begin{table}[t]\scriptsize
\begin{center}
\begin{tabular}{c||c|c|c|c}
\hline
 & EDVR$_{small}$ & Model 1 & Model 2 &  VESR-Net$_{small}$ \\
\hline
Separate NL  &       & \checkmark &            & \checkmark \\
CARB         &       &            & \checkmark & \checkmark  \\
Network Size & Small & Small      & Small      & Small       \\
\hline
PSNR  & 35.47 & 35.75 & 35.70 & {\bf 35.88}  \\
FLOPs & 152.93G & 154.83G & 152.99G & 162.14G  \\
Params & 14.80M & 14.86M & 15.00M & 15.96M  \\
\hline
\end{tabular}
\end{center}
\caption{The ablation study on Separate Non-Local (Separate NL) and channel-attention residual block (CARB). FLOPs \cite{molchanov2016pruning} are calculated on low-resolution video frames with the size of 64 $\times$ 64.}
\label{ablation}
\end{table}

\subsection{Evaluation Results}
As shown in Table \ref{experiment}, our baseline is EDVR$_{small}$. After increasing the number of CARB block to 49, EDVR achieves 0.28 dB gain in terms of PSNR compared with EDVR$_{small}$. The FLOPs and parameters of EDVR are remarkably larger than VESR-Net$_{small}$, but VESR-Net$_{small}$ can outperform EDVR with less computation complexity. With the help of our designed Seperal NL and CARB, VESR-Net is 0.22dB better than EDVR, which demonstrates the efficiency of VESR-Net. 

To evaluate the effectiveness of Separate Non-Local (Separate NL) operation and channel attention residual block (CARB) respectively, we conduct ablation studies on VESR-Net$_{small}$ and also compare with EDVR$_{small}$ on the dataset used in the second phase of Youku-VESR challenge. As shown in Table~\ref{ablation}, compared to EDVR$_{small}$, Model 1 introduces the Separate NL based on EDVR$_{small}$, achieving an improvement of 0.28db without increasing much parameters and FLOPs. With our CARB, Model 2 is nearly 0.23dB better than EDVR$_{small}$ with roughly the same model parameters and computational cost, demonstrating the effectiveness of CARB. At last, VESR-Net$_{small}$ introduces both Separate NL and CARB and achieves 0.41 dB, 0.13 dB and 0.18 dB performance gain compared to EDVR$_{small}$, Model 1 and Model 2 respectively. 

We further show some qualitative results of EDVR and VESR-Net on the dataset used in Youku-VESR challenge. As shown in Figure~\ref{qualitative_res}, VESR-Net can clearly restore the detail of subtitle in video frames.

\begin{table}[t]
\small
\begin{center}
\begin{tabular}{c||c}
\hline
Team & Score  \\
\hline
Avengers Assemble  & 37.851  \\
NJU$\_$L1         &   37.681  \\
ALONG$\_$NTES     & 37.632  \\
\hline
\end{tabular}
\end{center}
\caption{The leaderborad results of Youku-VESR challenge, where ``Avengers Assemble'' is our team while ``NJU\_L1'' and ``ALONG\_NTES'' are the teams ranking in the second and third place.}
\label{challenge_res}
\end{table}

\subsection{The Leaderborad Results for Youku-VESR Challenge}
Among 1500 registered teams in challenge, 10 teams entered in the final phase and submitted results, codes/executables, and factsheets. We list the score of top 3 teams in Table~\ref{challenge_res}. The score is computed with the following formula: \textit{PSNR score} $\times 80 \% +$\textit{VMAF score} $ \times 20\%$. Our team ``Avengers Assemble'' ranked the first place in the challenge, with about 0.2 point higher than the second and third teams.

\section{Conclusion}
In this paper, we proposed VESR-Net, an efficient network for video enhancement and super resolution. VESR-Net consists of a fusion module that leverages Separate Non-Local to capture the relation among video frames for frames fusion, and a reconstruction module that leverages CARB for efficient frame reconstruction. In Youku-VESR challenge, our proposed VESR-Net beat other competitive participants and ranked the first place.


{\small
\bibliographystyle{ieee_fullname}
\bibliography{egbib}

\begin{thebibliography}{10}\itemsep=-1pt

\bibitem{caballero2017real}
Jose Caballero, Christian Ledig, Andrew Aitken, Alejandro Acosta, Johannes
  Totz, Zehan Wang, and Wenzhe Shi.
\newblock Real-time video super-resolution with spatio-temporal networks and
  motion compensation.
\newblock In {\em Proceedings of the IEEE Conference on Computer Vision and
  Pattern Recognition}, pages 4778--4787, 2017.

\bibitem{devlin2018bert}
Jacob Devlin, Ming-Wei Chang, Kenton Lee, and Kristina Toutanova.
\newblock Bert: Pre-training of deep bidirectional transformers for language
  understanding.
\newblock {\em arXiv preprint arXiv:1810.04805}, 2018.

\bibitem{gavade2014super}
A Gavade and P Sane.
\newblock Super resolution image reconstruction by using bicubic interpolation.
\newblock In {\em National Conference on Advanced Technologies in Electrical
  and Electronic Systems}, volume~10, 2014.

\bibitem{haris2019recurrent}
Muhammad Haris, Gregory Shakhnarovich, and Norimichi Ukita.
\newblock Recurrent back-projection network for video super-resolution.
\newblock In {\em Proceedings of the IEEE Conference on Computer Vision and
  Pattern Recognition}, pages 3897--3906, 2019.

\bibitem{hyun2018spatio}
Tae Hyun~Kim, Mehdi~SM Sajjadi, Michael Hirsch, and Bernhard Scholkopf.
\newblock Spatio-temporal transformer network for video restoration.
\newblock In {\em Proceedings of the European Conference on Computer Vision
  (ECCV)}, pages 106--122, 2018.

\bibitem{jo2018deep}
Younghyun Jo, Seoung Wug~Oh, Jaeyeon Kang, and Seon Joo~Kim.
\newblock Deep video super-resolution network using dynamic upsampling filters
  without explicit motion compensation.
\newblock In {\em Proceedings of the IEEE Conference on Computer Vision and
  Pattern Recognition}, pages 3224--3232, 2018.

\bibitem{kim20183dsrnet}
Soo~Ye Kim, Jeongyeon Lim, Taeyoung Na, and Munchurl Kim.
\newblock 3dsrnet: Video super-resolution using 3d convolutional neural
  networks.
\newblock {\em arXiv preprint arXiv:1812.09079}, 2018.

\bibitem{kingma2014adam}
Diederik~P Kingma and Jimmy Ba.
\newblock Adam: A method for stochastic optimization.
\newblock {\em arXiv preprint arXiv:1412.6980}, 2014.

\bibitem{li2019fast}
Sheng Li, Fengxiang He, Bo Du, Lefei Zhang, Yonghao Xu, and Dacheng Tao.
\newblock Fast spatio-temporal residual network for video super-resolution.
\newblock {\em arXiv preprint arXiv:1904.02870}, 2019.

\bibitem{li2016toward}
Zhi Li, Anne Aaron, Ioannis Katsavounidis, Anush Moorthy, and Megha Manohara.
\newblock Toward a practical perceptual video quality metric.
\newblock {\em The Netflix Tech Blog}, 6, 2016.

\bibitem{liu2013bayesian}
Ce Liu and Deqing Sun.
\newblock On bayesian adaptive video super resolution.
\newblock {\em IEEE transactions on pattern analysis and machine intelligence},
  36(2):346--360, 2013.

\bibitem{molchanov2016pruning}
Pavlo Molchanov, Stephen Tyree, Tero Karras, Timo Aila, and Jan Kautz.
\newblock Pruning convolutional neural networks for resource efficient
  inference.
\newblock {\em arXiv preprint arXiv:1611.06440}, 2016.

\bibitem{nah2019ntire}
Seungjun Nah, Sungyong Baik, Seokil Hong, Gyeongsik Moon, Sanghyun Son, Radu
  Timofte, and Kyoung Mu~Lee.
\newblock Ntire 2019 challenge on video deblurring and super-resolution:
  Dataset and study.
\newblock In {\em Proceedings of the IEEE Conference on Computer Vision and
  Pattern Recognition Workshops}, pages 0--0, 2019.

\bibitem{sajjadi2018frame}
Mehdi~SM Sajjadi, Raviteja Vemulapalli, and Matthew Brown.
\newblock Frame-recurrent video super-resolution.
\newblock In {\em Proceedings of the IEEE Conference on Computer Vision and
  Pattern Recognition}, pages 6626--6634, 2018.

\bibitem{shi2016real}
Wenzhe Shi, Jose Caballero, Ferenc Husz{\'a}r, Johannes Totz, Andrew~P Aitken,
  Rob Bishop, Daniel Rueckert, and Zehan Wang.
\newblock Real-time single image and video super-resolution using an efficient
  sub-pixel convolutional neural network.
\newblock In {\em Proceedings of the IEEE conference on computer vision and
  pattern recognition}, pages 1874--1883, 2016.

\bibitem{stuhlmuller2000analysis}
Klaus Stuhlmuller, Niko Farber, Michael Link, and Bernd Girod.
\newblock Analysis of video transmission over lossy channels.
\newblock {\em IEEE Journal on Selected Areas in communications},
  18(6):1012--1032, 2000.

\bibitem{tao2017detail}
Xin Tao, Hongyun Gao, Renjie Liao, Jue Wang, and Jiaya Jia.
\newblock Detail-revealing deep video super-resolution.
\newblock In {\em Proceedings of the IEEE International Conference on Computer
  Vision}, pages 4472--4480, 2017.

\bibitem{vaswani2017attention}
Ashish Vaswani, Noam Shazeer, Niki Parmar, Jakob Uszkoreit, Llion Jones,
  Aidan~N Gomez, {\L}ukasz Kaiser, and Illia Polosukhin.
\newblock Attention is all you need.
\newblock In {\em Advances in neural information processing systems}, pages
  5998--6008, 2017.

\bibitem{wang2019edvr}
Xintao Wang, Kelvin~CK Chan, Ke Yu, Chao Dong, and Chen Change~Loy.
\newblock Edvr: Video restoration with enhanced deformable convolutional
  networks.
\newblock In {\em Proceedings of the IEEE Conference on Computer Vision and
  Pattern Recognition Workshops}, pages 0--0, 2019.

\bibitem{wang2018non}
Xiaolong Wang, Ross Girshick, Abhinav Gupta, and Kaiming He.
\newblock Non-local neural networks.
\newblock In {\em Proceedings of the IEEE Conference on Computer Vision and
  Pattern Recognition}, pages 7794--7803, 2018.

\bibitem{wiegand2003overview}
Thomas Wiegand, Gary~J Sullivan, Gisle Bjontegaard, and Ajay Luthra.
\newblock Overview of the h. 264/avc video coding standard.
\newblock {\em IEEE Transactions on circuits and systems for video technology},
  13(7):560--576, 2003.

\bibitem{xue2019video}
Tianfan Xue, Baian Chen, Jiajun Wu, Donglai Wei, and William~T Freeman.
\newblock Video enhancement with task-oriented flow.
\newblock {\em International Journal of Computer Vision}, 127(8):1106--1125,
  2019.

\bibitem{zhang2018image}
Yulun Zhang, Kunpeng Li, Kai Li, Lichen Wang, Bineng Zhong, and Yun Fu.
\newblock Image super-resolution using very deep residual channel attention
  networks.
\newblock In {\em Proceedings of the European Conference on Computer Vision
  (ECCV)}, pages 286--301, 2018.

\end{thebibliography}
}

\end{document}